\documentclass[11pt,a4paper]{article}
\usepackage[hyperref]{AACL-IJCNLP2020}
\usepackage{times}
\usepackage{latexsym}

\usepackage{microtype}

\aclfinalcopy 

\usepackage{graphicx}
\usepackage{booktabs}
\usepackage{amsmath}
\usepackage{amssymb}
\usepackage{url}
\usepackage{adjustbox}
\usepackage{caption}
\usepackage{natbib}
\usepackage{subcaption}
\usepackage{comment}
\usepackage{url}
\usepackage{multirow,array}
\usepackage{xspace}
\usepackage{pifont}
\usepackage{footmisc}

\definecolor{cadmiumgreen}{rgb}{0.0, 0.42, 0.24}
\definecolor{brightmaroon}{rgb}{0.86, 0.078, 0.23}

\newcommand{\cmark}{\textcolor{cadmiumgreen}{\ding{51}}}%
\newcommand{\xmark}{\textcolor{red}{\ding{55}}}

\DeclareMathOperator*{\argmin}{arg\,min}

\newcommand{\p}{\textsc{Penman}\xspace}

\title{AMR Quality Rating with a Lightweight CNN}

\author{Juri Opitz\\
 Dept.\ of Computational Linguistics \\ 
 Heidelberg University \\
 69120 Heidelberg\\
 \texttt{opitz@cl.uni-heidelberg.de}}

\date{}

\begin{document}

\maketitle

\begin{abstract}
Structured semantic sentence representations such as Abstract Meaning Representations (AMRs) are potentially useful in various NLP tasks. However, the quality of automatic parses can vary greatly and jeopardizes their usefulness. This can be mitigated by models that can accurately rate AMR quality in the absence of costly gold data, allowing us to inform downstream systems about an incorporated parse's trustworthiness or select among different candidate parses.

In this work, we propose to transfer the AMR graph to the domain of images. This allows us to create a simple convolutional neural network (CNN) that imitates a human judge tasked with rating graph quality. Our experiments show that the method can rate quality more accurately than strong baselines, in several quality dimensions. Moreover, the method proves to be efficient and reduces the incurred energy consumption.
\end{abstract}

\section{Introduction}

\begin{figure}
    \centering
    \includegraphics[width=0.78\linewidth]{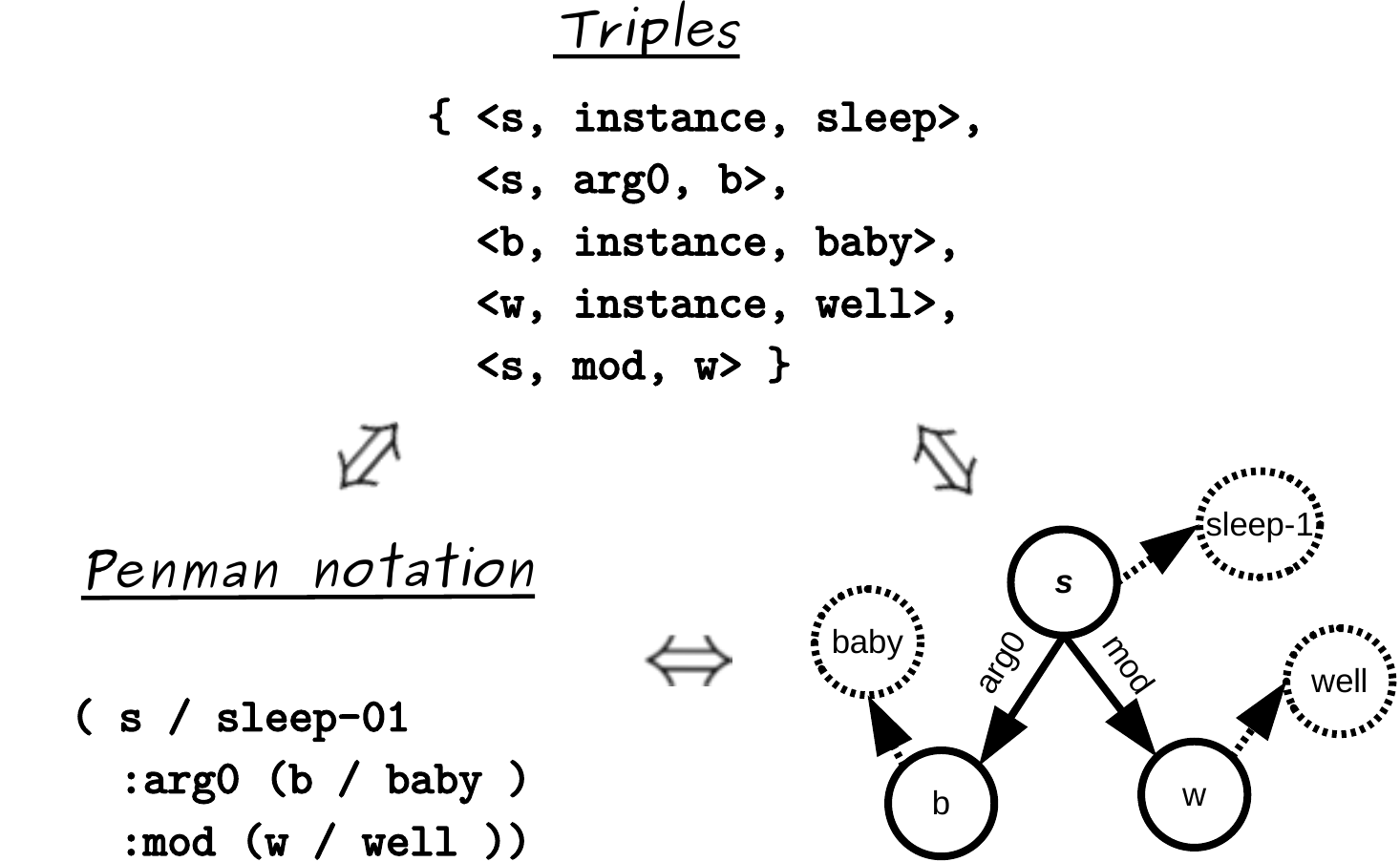}
    \caption{Equivalent representations of the AMR for \textit{``The baby is sleeping well''.} 
    }
    \label{fig:ex1}
\end{figure}

The goal of sentence meaning representations is to capture the meaning of sentences in a well-defined format. One of the most prominent frameworks for achieving this is \textit{Abstract Meaning Representation} (AMR) \citep{banarescu2013abstract}. In AMR, sentences are represented as directed acyclic and rooted graphs. An example is displayed in Figure \ref{fig:ex1}, where we see three equivalent displays of an AMR that represents the meaning of the sentence \textit{``The baby is sleeping well''}. In AMR, nodes are variables or concepts, while (labeled) edges express their relations. Among other phenomena, this allows AMR to capture coreference (via re-entrant structures) and semantic roles (via \textit{:arg}$_n$ relation). Furthermore, AMR links sentences to KBs: e.g., predicates are mapped to PropBank \citep{palmer2005proposition,kingsbury-palmer-2002-treebank}, while named entities are linked to Wikipedia. From a logical perspective, AMR is closely related to first-order logic (FOL, see \citet{bos-first-order,bos2019separating} for translation mechanisms).


\begin{figure}
    \centering
    \includegraphics[width=0.6\linewidth]{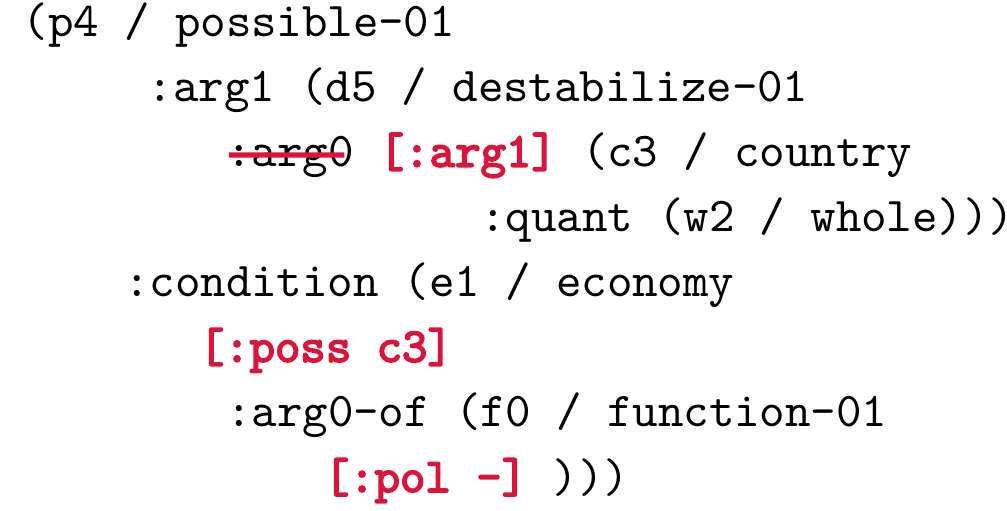}
    \caption{Parse of \textit{Without a functioning economy, the whole country may destabilize} with \color{brightmaroon}{errors} outlined. 
    }
    \label{fig:exparse}
\end{figure}

Currently, AMRs are leveraged to enhance a variety of natural language understanding tasks. E.g., they have enhanced commonsense reasoning and question answering \citep{AAAI1612345}, machine translation \citep{song-etal-2019-semantic}, text summarization \citep{liao-etal-2018-abstract,dohare2017text} and paraphrasing \citep{issa-etal-2018-abstract}. However, there is a critical issue with automatically generated AMRs (parses): they are often deficient.

These deficiencies can be quite severe, 
even when high-performance parsers are used. For example, in Figure \ref{fig:exparse}, a neural parser  \citep{lyu-titov-2018-amr} conducts several errors when parsing \textit{Without a functioning economy the whole country may destabilize}. E.g., it misses a negative polarity and classifies a patient argument as the agent by failing to see that \textit{destabilize} here functions as an ergative verb (parser: the country is the causer of destabilize; correct: the country is the object that is destabilized). In sum, the parse has misrepresented the sentence's meaning.\footnote{\textit{?With a functioning economy, the whole country may cause something to destabilize.}} However, assessing such deficiencies via comparison against a gold reference (as in classical parser evaluation) is often infeasible in practice: it takes a trained annotator and appr.\ 10 minutes to manually create one AMR \citep{banarescu2013abstract}.

To mitigate these issues, we would like to automatically rate the quality of AMRs without the costly gold graphs. This would allow us to signal downstream task systems the incorporated graphs' trustworthiness or select among different candidate graphs from different parsing systems. To achieve this, we propose a method that imitates a human rater, who is inspecting the graphs. We show that the method can efficiently rate the quality of the AMRs in the absence of gold graphs.

The remainder of the paper is structured as follows: in Section \ref{sec:idea}, we outline our idea to exploit the textual multi-line string representation of AMRs, allowing for efficient and simple AMR processing while preserving  vital graph structure. In Section \ref{sec:inst}, we instantiate this idea in a lightweight CNN that predicts the quality of AMR graphs along multiple dimensions of interest. In our experiments (Section \ref{sec:exp}), we show that this framework is efficient and performs better than strong baselines. 
Our code is available at \url{https://github.com/flipz357/amr-quality-rater}.

\section{AMR as image with latent channels}\label{sec:idea}

In this section, we first motivate to treat AMRs as images with latent channels in order to rate them efficiently. 
Second, we briefly describe the task at hand: Rating the quality of AMR graphs in the absence of gold graphs. Finally, to solve this task, we create a lightweight CNN that evaluates AMR quality in multiple dimensions of interest.

\begin{figure}
\includegraphics[width=\linewidth]{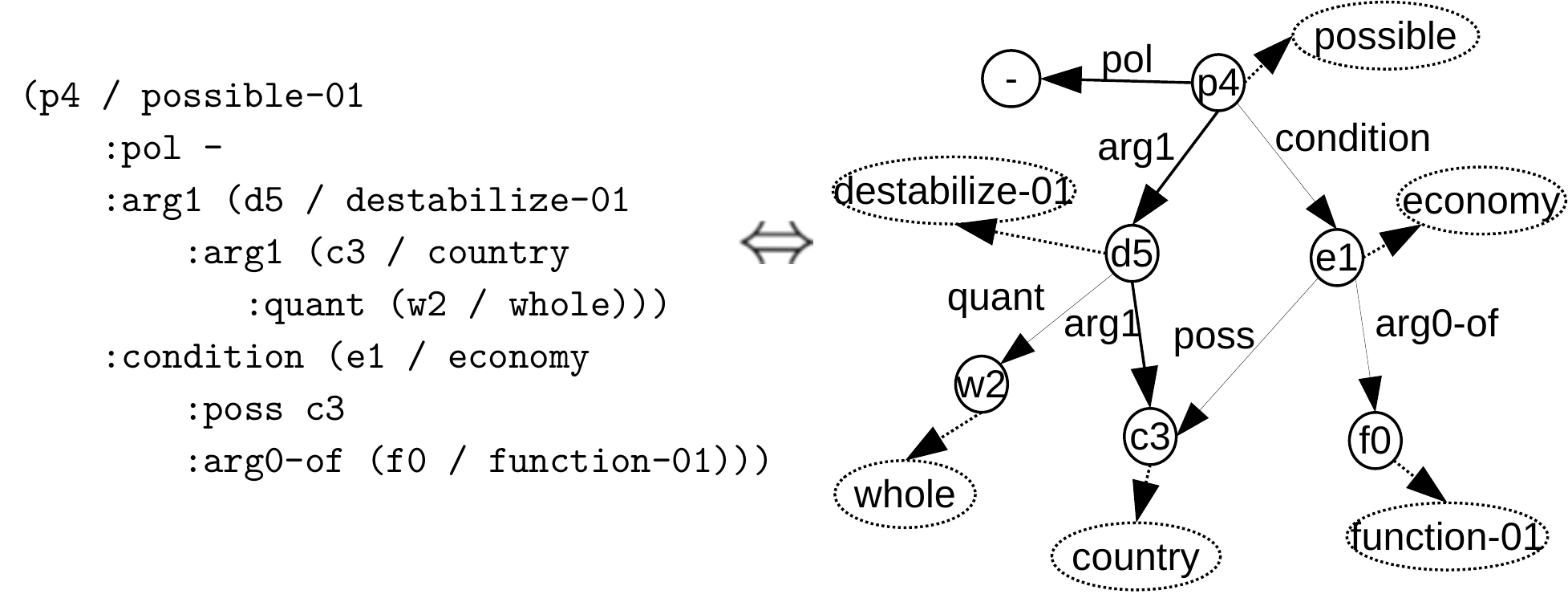}
\caption{Different displays for an AMR structure of a sentence that has medium length (left: \p notation, right: graphical visualization)}
\label{sfig:medsize}
\end{figure}

\begin{table}
    \centering
    \scalebox{0.53}{
    \begin{tabular}{|l|lll|}
    \toprule
       graph representation  & computer processing & human understanding & well-defined \\
         \midrule
         
triples; $G=(V, E, g, f, \Sigma)$  & \cmark (e.g., GNN) & \xmark & \cmark  \\
graph visualization  & \xmark & \cmark~(short sentences) & \xmark \\
\p, linearized string  & \cmark (e.g., LSTM)& \xmark & \cmark \\
\p, indents & \cmark (this work) & \cmark & \cmark\\
\bottomrule
    \end{tabular}}
    \caption{Equivalent AMR representations and their accessibility with respect to human or computer (\cmark: `okay', \xmark: `perhaps possible, but difficult'). $(V, E, f, g, \Sigma)$: $V$ and $E$ are sets of vertices and edges, $g,f$ assign node and edge labels from a vocabulary $\Sigma$.}
    \label{tab:adv}
\end{table}
\paragraph{The \p notation and its (hidden) advantages} The native AMR notation is called \p-notation or \textit{Sentence Plan Language} \citep{Kasper:1989:FIL:100964.100979,Mann:1983:OPT:2886844.2886899}. 
 Provably, an advantage of this notation is that it allows for secure AMR storage in text files. However, we argue that it has more advantages. For example, due to its clear structure, it allows humans a fairly quick understanding even of medium-sized to large AMR structures (Figure \ref{sfig:medsize}, left). On the other hand, we argue that a graphical visualization of such medium-sized to large AMRs (Figure \ref{sfig:medsize}, right) could hamper intuitive understanding, since the abundant visual signals (circles, arrows, etc.) may more easily overwhelm humans. Moreover, in every display, one would depend on an algorithm that needs to determine a suitable (and spacious) arrangement of the nodes, edges and edge labels. It may be for these reasons, that in the AMR annotation tool\footnote{\url{https://www.isi.edu/cgi-bin/div3/mt/amr-editor/login-gen-v1.7.cgi}}, the graph that is under construction is always shown in \p notation to the human user. 

In sum, we find that the indented multi-line \p form possesses three key advantages (Table \ref{tab:adv}): (i) it enables fairly easy human understanding, (ii), it is well-defined and (iii), which is what we will show next, it can be computationally exploited to better rate AMR quality.  

\begin{figure}
    \centering
    \includegraphics[width=1.0\linewidth]{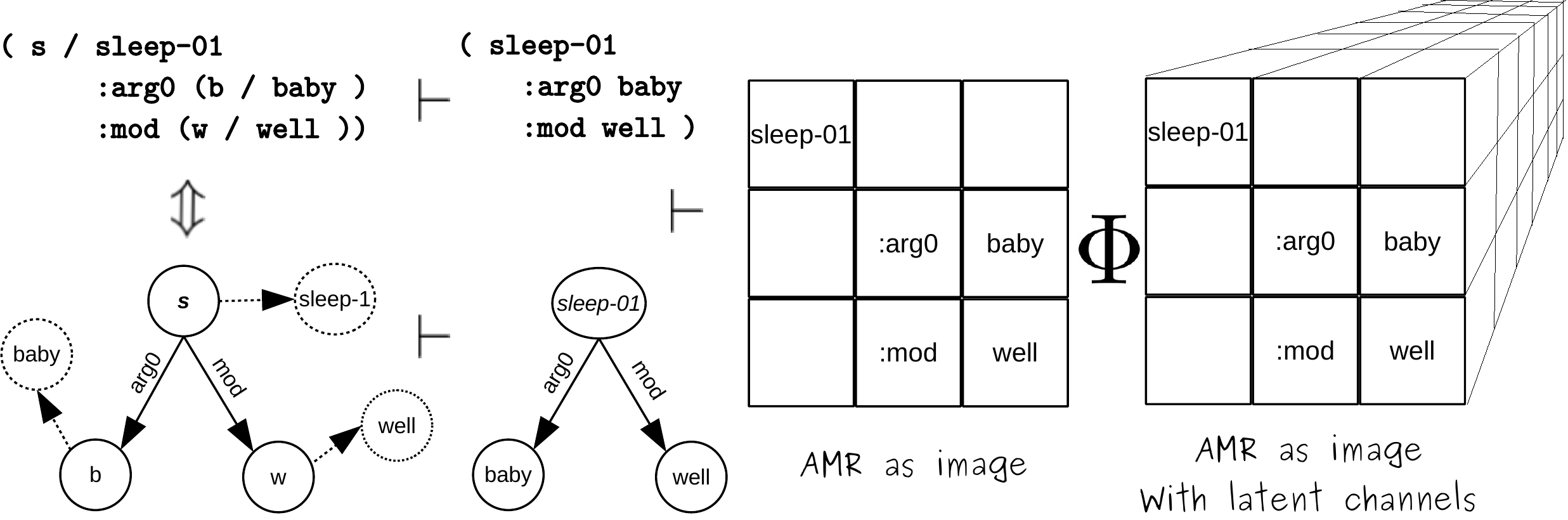}
    \caption{We transform the (simplified) \textsc{Penman} representation to an image and use $\Phi$ to add latent channels.
    }
    \label{fig:latentdims}
\end{figure}

\paragraph{AMR as image to preserve graph structure} Figure \ref{fig:latentdims} describes our proposed sentence representation treatment. 
After non-degenerate AMR graph simplification (more details in \textit{Preprocessing}, \ref{par:prepro}) 
, we first project the \p representation onto a small grid (`image'). Each AMR token (e.g., a node or an edge) is represented as a `categorical pixel'. Second, $\Phi$ adds latent `channels' to the categorical pixels, which can be learned incrementally in an application.  In other words, every AMR token is represented by a fixed-sized vector of real numbers. These vectors are arranged such that the original graph structure is fully preserved. 

\subsection{Task: Rating the quality of AMR graphs} We aim at rating the quality of AMR graphs (`parses') in the absence of gold graphs. This boils down to answering the following question: \textit{how well does a candidate AMR graph capture a given natural language sentence}? Therefore, the exact goal in this task is to learn a mapping

\begin{equation}\label{eq:1}
    f: \mathcal{S} \times \mathcal{G} \rightarrow \mathbb{R}^d, 
\end{equation}

that maps a sentence $s \in \mathcal{S}$ together with a candidate AMR graph $g \in \mathcal{G}$ onto $d$ scores, which describe the AMR with regard to $d$ quality dimensions of interest. A successful mapping function should strongly correlate with the gold scores as they would emerge from evaluation against gold graphs. We proceed by describing the targeted dimensions in more detail.

\paragraph{Main AMR quality dimensions} The main quality dimensions that we desire our model to predict are estimated \textbf{Smatch F1}/\textbf{recall}/\textbf{precision}. Smatch is the canonical AMR metric, assessing the triple overlap between two graphs, after an alignment step \citep{cai-knight-2013-smatch}. 

\paragraph{AMR sub-task quality dimensions} \label{par:qdims} However, we predict also other quality dimensions to assess various AMR aspects \citep{damonte-etal-2017-incremental}. In this place, we can merely provide a brief overview: (i) \textbf{Unlabeled}: Smatch F1 when disregarding edge-labels. (ii) \textbf{No WSD}: Smatch F1  when ignoring ProbBank senses. (iii) \textbf{Frames}: PropBank frame identification F1 (iii) \textbf{Wikification}: KB linking F1 score on \textit{:wiki} relations. (iv) \textbf{Negations}: negation detection F1. (v) \textbf{NamedEnt}: NER F1. (vi) \textbf{NS frames}: F1 score for ProbBank frame identification when disregarding the sense. (vii) \textbf{Concepts} F score for concept identification (viii) \textbf{SRL}: Smatch F1 computed on arg-i roles only. (ix) \textbf{Reentrancy}: Smatch F1 computed on re-entrant edges only. (x) \textbf{IgnoreVars}: F1 when variable nodes are ignored. (xi) \textbf{Concepts}: F1 for concept detection.

\subsection{A lightweight CNN to rate AMR quality}
\label{sec:inst}
\begin{figure*}
    \centering
    \includegraphics[trim=0cm 0cm 0cm 0cm,clip,width=\linewidth]{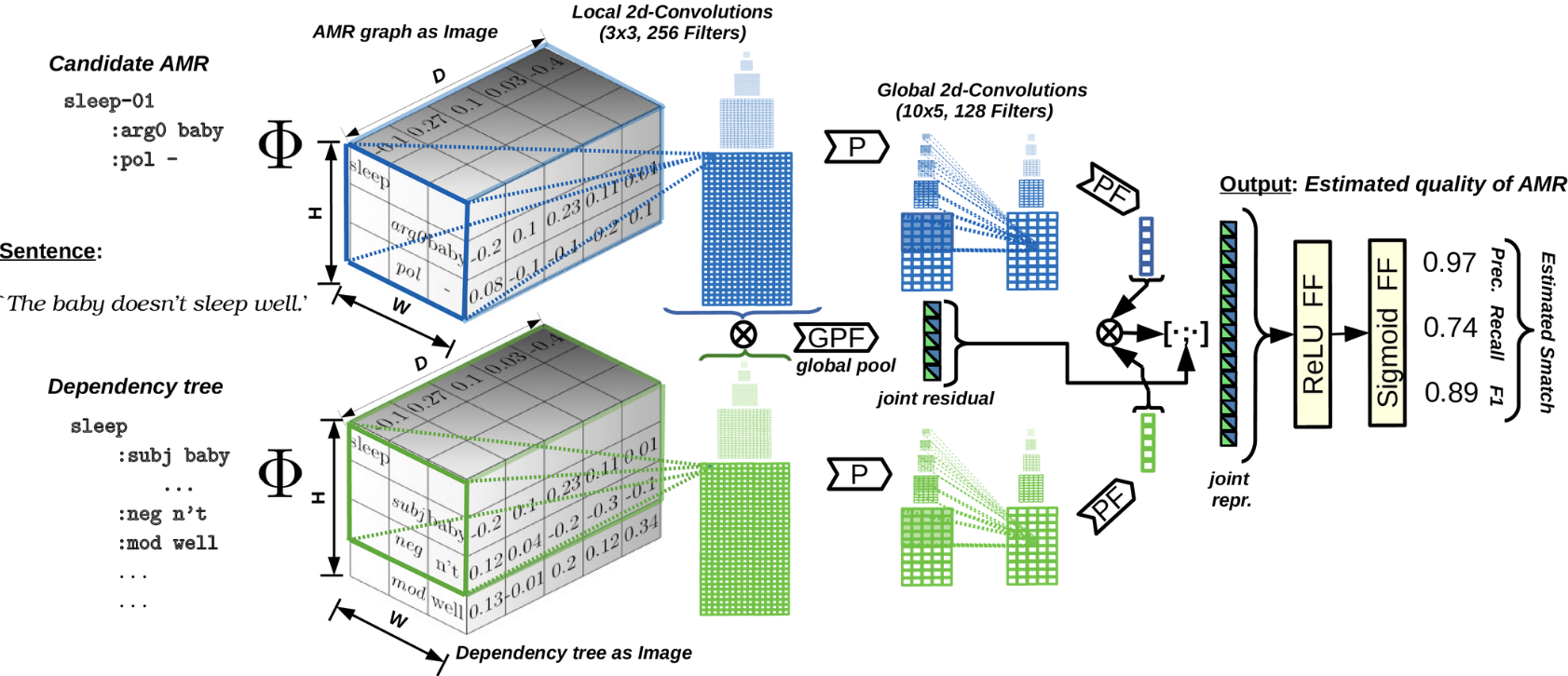}
    \caption{Our proposed architecture for efficient AMR quality assessment.}
    \label{fig:arch}
\end{figure*}

\label{par:model} We want to model $f$ (Eq. \ref{eq:1}) in order to estimate a suite of quality scores $y \in \mathbb{R}^d$ for any automatically generated AMR graph, given only the graph and the sentence from whence it is derived. Following \citet{opitz-frank-2019-automatic}, we will contrast the AMR against the sentence's dependency parse, exploiting observed structural similarities between these two types of information \cite{wang-etal-2015-transition}. Our proposed method allows this in a simple way by processing dependency and AMR graphs in parallel. The architecture is outlined in Figure \ref{fig:arch}. 

\paragraph{Symbol embedding} The latent channels of AMR and dependency `pixels' represent the embeddings of the `tokens' or `symbols' contained in the AMR and dependency vocabulary. These symbols represent nodes or edges. We use two special tokens: the \texttt{<tab>} token, which represents the indention level, and the \texttt{<pad>} token, which fills the remaining empty `pixels'. 
By embedding lookup, we obtain AMR and dependency images with 128 latent channels and 45x15 `pixels' ($\Phi$ in Figure \ref{fig:arch}; the amount of pixels is chosen such that more than 95\% of training AMRs can be fully captured).

\paragraph{Encoding local graph regions} Given AMR and dependency images with 128 latent channels and 45x15 pixels, we apply to each of the two images 256 filters of size 3x3, which is a standard type of kernel in CNNs. This converts both graphs to 256 feature maps each $\in \mathbb{R}^{45\times 15}$ (same-padding), obtaining two three-dimensional tensors $L^1_{amr}, L^1_{dep} \in \mathbb{R}^{45\times 15 \times 256}$. From here, we construct our first joint representation, which matches local dependency regions with local AMR regions:
\begin{equation}
    j_{res} = GPF(L^1_{amr} \otimes L^1_{dep}),
\end{equation}

where $x\otimes y = [x\odot y;x\ominus y]$ denotes the concatenation of element-wise multiplication and element-wise subtraction. $GPF$ is an operation that performs global pooling and vectorization (`flattening') of any input tensor. This means that $j_{res} \in \mathbb{R}^{512}$ is a joint representation of the locally matched dependency and AMR graph regions. This intermediate process is outlined in Figure \ref{fig:arch} by $\otimes$ (left) and GPF. Finally, we reduce the dimensions of the two intermediate three-dimensional representations $L^1_{amr}$ and $L^1_{dep}$ with 3x3 max-pooling and obtain $L^2_{amr}$ and $L^2_{dep}$ $\in \mathbb{R}^{15\times 5 \times 256}$

\paragraph{Encoding global graph regions} For a moment, we put the \textit{joint residual} ($j_{res}$) aside and proceed by processing the locally convolved feature maps with larger filters. While the first convolutions allowed us to obtain abstract \textit{local} graph regions $L^2_{amr}$ and $L^2_{dep}$, we now aim at matching more \textit{global} regions. More precisely, we use 128 2D filters of shape 10x5, followed by a 5x5 max-pooling operations on $L^2_{amr}$ and $L^2_{dep}$. Thus, we have obtained vectorized abstract global graph representations $g_{amr}, g_{dep} \in \mathbb{R}^{384}$. Then, we construct a joint representation (right $\otimes$, Figure \ref{fig:arch}):
\begin{equation}
j_{glob} = g_{amr} \otimes g_{dep}.
\end{equation}

At this point, together with the joint residual representation from the local region matching, we have arrived at two joint vector representations $j_{glob}$ and $j_{res}$. We concatenate  them ($[\cdot;\cdot]$ in Figure \ref{fig:arch}) to form one joint representation $\mathbf{j} \in \mathbb{R}^{1280}$:

\begin{equation}
    \mathbf{j} = [j_{res};j_{glob}]
\end{equation}

\paragraph{Quality prediction} The shared representation $\mathbf{j}$ is further processed by a feed-forward layer with ReLU activation functions ($FF_{+ReLU}$, Figure \ref{fig:arch}) and a consecutive feed-forward layer with sigmoid activation functions ($FF_{+sigm}$, Figure \ref{fig:arch}):
\begin{equation}
    \mathbf{y}=sigm(ReLU(\mathbf{j}^TA)B),
\end{equation}
where $A \in \mathbb{R}^{1280 \times h}$, $B \in \mathbb{R}^{h \times dim(out)}$ are parameters of the model and $sigm(x) = (\frac{1}{1+e^{-x_1}},...,\frac{1}{1+e^{-x_{dim(out)}}})$ projects $x$ onto  $[0,1]^{dim(out)}$. When estimating the main AMR metric scores we instantiate three output neurons ($dim(out)=3$) that represent estimated Smatch precision, Smatch recall and Smatch F1. In the case where we are interested in a more fine-grained assessment of AMR quality (e.g., knowledge-base linking quality), we have 33 output neurons representing expected scores for various semantic aspects involved in AMR parsing (we predict precision, recall and F1 of 11 aspects, as  outlined in \S \ref{par:qdims}). 

To summarize, the residual joint representation should capture local similarities. On the other hand, the second joint representation aims to capture the more global and structural properties of the two graphs. Both types of information inform the final quality assessment of our model in the last layer.

\section{Experiments}\label{sec:exp}

In this section, we first describe the data, changes to the data that target the reduction of biases, and the baseline. After discussing our main results, we conduct further analyses. (i), we study the effects of our data-debiasing steps. (ii), we assess the performance of our model in a classification task (distinguishing good from bad parses). (iii), we assess the model performance when we only provide the candidate AMR and the sentence (dependency tree ablation). (iv), we provide detailed measurements of the method's computational cost.

\subsection{Experimental setup}

\paragraph{Data} We use the data from \citet{opitz-frank-2019-automatic}. The data set consists of more than 15,000 sentences with more than  60,000 corresponding parses, by three different automatic parsing systems and a human. More precisely, the data set $\mathcal{D}= \{(s_i,g_i,y_i)\}_{i=1}^N$ consists of tuples $(s_i,g_i,y_i)$, where $s_i \in \mathcal{S}$ is a natural language sentence, $g_i \in \mathcal{G}$ is a `candidate' AMR graph and $y_i \in \mathbb{R}^d$ is a 36-dimensional vector containing scores which represent the quality of the AMR graph in terms of precision, recall and F1 with respect to 12 different tasks captured by AMR (as outlined in \S \ref{par:qdims}).

\paragraph{Debiasing of the data} We observe three biases in the data. First, the graphs in the training section of our data are less deficient than in the development and testing data, because the parsers were trained on \textit{(sentence, gold graph)} pairs from the training section. For our task, this means that the training section's target scores are higher, on average, than the target scores in the other data partitions. To achieve more balance in this regard, we re-split the data randomly on the sentence-id level (such that a sentence does not appear in more than one partition with different parses). 

Second, we observe that the data contains some superficial hidden clues that could give away the parse's source. This bears the danger that a model does not learn to assess the \textit{parse quality}, but to assess the \textit{source of the parse}. And since some parsers are better or worse than others, the model could exploit this bias. For example, consider that one parser prefers to write \textit{(r / run-01 :arg1 (c / cat) :polarity - )}, while the other parser prefers to write \textit{(r / run-01 :polarity - :arg1 (c / cat) )}. These two structures are semantically equivalent but differ on the surface. Hence, the arrangement of the output may provide unwanted clues on the source of the parse. To alleviate this issue, we randomly re-arrange all parses on the surface, keeping their semantics.\footnote{Technically, this is achieved by reformatting the parses such that in the depth-first writing-traversal at node $n$ the out-going edges of $n$ will be traversed in random order.}\footnote{Different variable names, e.g., \textit{(r / run-01) and (x / run-01 )} are not an issue in this work since the variables are handled via \cite{van-noord-bos-2017-dealing}. See also \textit{Preprocessing}, \S \ref{par:prepro}}

A third bias stems from a design choice in the metric scripts used to calculate the target scores. More precisely, the extended $Smatch$-metric script, per default, assigns a parse that does not contain a certain edge-type (e.g., :arg$_{n}$) the score 0 with respect to the specific quality dimension (in this case, SRL: 0.00 Precision/Recall/F1). However, if the gold parse also does not contain an edge of this type (i.e., :arg$_{n}$), then we believe that the correct default score should be 1, since the parse is, in the specific dimension, in perfect agreement with the gold (i.e., SRL: 1.00 Precision/Recall/F1). Therefore, we set all sub-task scores, where the predicted graph agrees with the gold graph in the absence of a feature, from 0 to 1.

\paragraph{Preprocessing}\label{par:prepro} Same as prior work, we dependency-parse and tokenize the sentences with spacy \citep{spacy2} and replace variables with corresponding concepts (e.g., \textit{(j / jump-01 :arg0 (g / girl))} is translated to \textit{(jump-01 :arg0 (girl))}. Re-entrancies are handled with pointers according to \citet{van-noord-bos-2017-dealing}, which ensures non-degenerate AMR simplification.\footnote{For example, consider the sentence \textit{The cat scratches itself} and its graph \textit{(x / scratch-01 :arg0 (y / cat) :arg1 y))}. Replacing the variables with concepts would come at the cost of an information loss w.r.t.\ to coreference: \textit{(scratch-01 :arg0 cat :arg1 cat)} --- does the cat scratch itself or another cat? Hence, pointers are used to translate the graph into \textit{(scratch-01 :arg0 *0* cat :arg1 *0*))}.} Furthermore, we lower-case all tokens, remove quotation marks and join sub-structures that represent names.\footnote{E.g.,  \textit{:name (name :op1 
`Barack' :op2 `Obama')} is translated to \textit{:name barack obama}.} The vocabulary encompasses all tokens of frequency $\geq$ 5, remaining ones are set to \texttt{<unk>}.

\paragraph{Training} 
All parameters are initialized randomly. We train for 5 epochs and select the parameters $\theta$ from the epoch where maximum development scores were achieved (with respect to average Pearson's $\rho$ over the quality dimensions). In training, we reduce the squared error with gradient descent (Adam rule \citep{kingma2014adam}, learning rate = 0.001, mini batch size = 64):
\vspace{-2mm}
\begin{equation}
    \theta^{*} = \argmin_{\theta} \sum_{i=1}^{|\mathcal{D}|} \sum_{j=1}^{|M|}(y_{i,j} - f_\theta(s_i,g_i)_j)^2,
\end{equation}
\vspace{-2mm}
where $M$ is the set of target metrics.

\paragraph{Baseline} Our main baseline is the model of previous work, henceforth denoted by \textsc{\textbf{LG-LSTM}}. The method works in the following steps: first, it uses a depth-first graph traversal to linearize the automatic AMR graph and the corresponding dependency tree of the sentence. Second, it constructs a joint representation and predicts the score estimations. To further improve its performance, the baseline uses some extra-features (e.g., a shallow alignment from dependency tokens to AMR tokens).\footnote{\label{fn:aux}Furthermore, the baseline uses auxiliary losses to achieve a slight performance gain in predicting the Smatch metrics. For the sake of simplicity, we do not use these auxiliary losses, except in one experiment, where we show that our method achieves a similar small gain with the auxiliary losses.} Generally speaking, the baseline is a model that works based on graph linearizations. Such type of model, despite its apparent simplicity, has proven to be an effective baseline or state-of-the-art method in various works about converting texts into graphs \citep{konstas-etal-2017-neural,van2017neural}, or converting graphs into texts \citep{bastings-etal-2017-graph,beck-etal-2018-graph,song19, pourdamghani-etal-2016-generating,song-etal-2018-graph, NIPS2015_5635,mager2020gpttoo}, or performing mathematically complex tasks modeled as graph-to-graph problems, such as symbolic integration \citep{Lample2020Deep}. However, in our main results, we also display the results of two additional baselines: \textbf{GNN} \cite{song-etal-2018-graph}, where we encode the dependency tree and the AMR with a graph-recurrent encoder and perform regression on the joint averaged node embedding vectors.\footnote{$\bigg[\frac{1}{|V_A|}\sum_{v \in V_{A}} emb(v)\bigg]\otimes\bigg[\frac{1}{|V_D|}\sum_{v \in V_{D}}emb(v)\bigg]$.} And \textbf{Ridge}, an l2-regularized linear regression that is based on shallow graph statistics.\footnote{For the dependency graph (D) and the AMR graph (A) we both compute $\phi(A|D)$ = [density, avg.\ node degree, node count, edge count, (arg0$|$subj) count, (arg1$|$obj) count], the final feature vector then is defined as $\Phi(x)$ = [$\phi$(A) - $\phi$(D); $\phi$(D); $\phi$(A); $\frac{|\text{lemmas(D)} \cap \text{concepts(A)}|}{|\text{lemmas(D)} \cup \text{concepts(A)}|}$]}

\subsection{Results}

\paragraph{Main AMR quality dimensions} The main quality of an AMR graph is estimated in expected triple match ratios (Smatch F1, Precision and Recall). 
\begin{table}
    \centering
   \scalebox{0.7}{\begin{tabular}{llrr|rrr}
    \toprule
       &Smatch  & Ridge & GNN & LG-LSTM & ours & change \% \\
       \midrule
       \parbox[t]{2mm}{\multirow{3}{*}{\rotatebox[origin=c]{90}{P's $\rho$}}} &F1 & 0.428 & 0.659 & 0.662$^{\pm 0.00}$ & 0.696$^{\pm 0.00}$ & +5.14 $\dagger\ddagger$\\ 
       &Precision & 0.348 & 0.601& 0.600$^{\pm 0.00}$ & 0.623$^{\pm 0.01}$ & +3.83 $\dagger~$\\
       & Recall & 0.463& 0.667 &0.676$^{\pm 0.00}$ & 0.719$^{\pm 0.00}$ & +6.36 $\dagger\ddagger$\\
       \midrule
       \midrule
    \parbox[t]{2mm}{\multirow{3}{*}{\rotatebox[origin=c]{90}{RMSE}}} &F1 & 0.155 & 0.132 & 0.130$^{\pm 0.00}$ & 0.128$^{\pm 0.00}$ & -1.54\\
       &Precision  &0.146 & 0.127 & 0.126$^{\pm 0.00}$ & 0.126$^{\pm 0.00}$ & +-0.0\\
       &Recall& 0.169 & 0.141 & 0.142$^{\pm 0.00}$ & 0.136$^{\pm 0.00}$ & -4.23\\
       \bottomrule
    \end{tabular}}
    \caption{Main results. Pearson's corr.\ coefficient (row 1-3) is better if higher; root mean square error (RMSE, row 4-6) is better if lower. The quality dimensions are explained in \S \ref{par:qdims}. 
    $\dagger$ ($\ddagger$): p $<$ 0.05 (p $<$ 0.005), significant difference in the correlations with two-tailed test using Fisher $\rho$ to z transformation \citep{fisher1915frequency}.}
    \label{tab:mainres}
\end{table}
The results, averaged over 10 runs, are displayed in Table \ref{tab:mainres}. With regard to estimated Smatch F1, we achieve a correlation with the gold scores of 0.695 Pearson's $\rho$. This constitutes a significant improvement of appr.\ 5\% over LG-LSTM. Similarly, recall and precision correlations improve  by 6.36\% and 3.83 \% (from 0.676 to 0.719 and 0.600 to 0.623). While the improvement in predicted recall is significant at p$<$0.05 and p$<$0.005, the improvement in predicted precision is significant at p$<$0.05. When we consider the root mean square error (RMSE), we find that the method improves over the best baseline by -1.54\%  in estimated Smatch F1 and -4.23\% in estimated Smatch recall. On the other hand, the RMS error in estimated precision remains unchanged. 

\paragraph{AMR subtask quality} Our model can also rate the quality of an AMR graph in a more fine-grained way.
\begin{table}
    \centering
    \scalebox{0.78}{
    \begin{tabular}{llrrr}
    \toprule
    &Quality Dim.\ & LG-LSTM & ours & change \% \\
    \midrule
 \parbox[t]{2mm}{\multirow{11}{*}{\rotatebox[origin=c]{90}{F1 Pearson's $\rho$}}} 
&Concepts & 0.508$^{\pm 0.01}$ & 0.545$^{\pm 0.01}$ & +7.28~$\dagger$~~\\
&Frames & 0.420$^{\pm 0.01}$ & 0.488$^{\pm 0.01}$ & +16.19~$\dagger\dagger$\\
&IgnoreVars & 0.627$^{\pm 0.01}$ & 0.665$^{\pm 0.00}$ & +6.06~$\dagger\dagger$\\
&NamedEnt. & 0.429$^{\pm 0.02}$ & 0.460$^{\pm 0.01}$ & +7.23~$\dagger$~~\\
&Negations & 0.685$^{\pm 0.02}$ & 0.746$^{\pm 0.01}$ & +8.91~$\dagger\dagger$\\
&NoWSD & 0.640$^{\pm 0.01}$ & 0.680$^{\pm 0.00}$ & +6.25~$\dagger\dagger$\\
&NS-frames & 0.419$^{\pm 0.02}$ & 0.505$^{\pm 0.01}$ & +20.53~$\dagger\dagger$\\
&Reentrancies & 0.508$^{\pm 0.01}$ & 0.602$^{\pm 0.00}$ & +18.50~$\dagger\dagger$\\
&SRL & 0.519$^{\pm 0.01}$ & 0.581$^{\pm 0.01}$ & +11.95~$\dagger\dagger$\\
&Unlabeled & 0.628$^{\pm 0.01}$ & 0.663$^{\pm 0.00}$ & +5.57~$\dagger\dagger$\\
&Wikification & 0.901$^{\pm 0.00}$ & 0.904$^{\pm 0.00}$ & +0.33~~~~~\\

\midrule
\midrule
\parbox[t]{2mm}{\multirow{11}{*}{\rotatebox[origin=c]{90}{F1 RMSE}}}
&Concepts & 0.117$^{\pm 0.00}$ & 0.114$^{\pm 0.00}$ & -2.56\\
&Frames & 0.186$^{\pm 0.00}$ & 0.182$^{\pm 0.00}$ & -2.15\\
&IgnoreVars & 0.195$^{\pm 0.00}$ & 0.186$^{\pm 0.00}$ & -4.62\\
&NamedEnt. & 0.159$^{\pm 0.00}$ & 0.156$^{\pm 0.00}$ & -1.89\\
&Negations & 0.197$^{\pm 0.00}$ & 0.180$^{\pm 0.00}$ & -8.63\\
&NoWSD & 0.132$^{\pm 0.00}$ & 0.126$^{\pm 0.00}$ & -4.55\\
&NS-frames & 0.157$^{\pm 0.00}$ & 0.155$^{\pm 0.00}$ & -1.27\\
&Reentrancies & 0.285$^{\pm 0.00}$ & 0.265$^{\pm 0.00}$ & -7.02\\
&SRL & 0.189$^{\pm 0.00}$ & 0.181$^{\pm 0.00}$ & -4.23\\
&Unlabeled & 0.124$^{\pm 0.00}$ & 0.121$^{\pm 0.00}$ & -2.42\\
&Wikification & 0.165$^{\pm 0.00}$ & 0.162$^{\pm 0.00}$ & -1.82\\

\bottomrule
    \end{tabular}}
    \caption{Results for AMR quality rating w.r.t.\ various sub-tasks. 
    $\dagger$ ($\ddagger$): significance (c.f.\ caption Table \ref{tab:mainres}).}
    \label{tab:subtaskres}
\end{table}
The results are displayed in Table \ref{tab:subtaskres}. Over almost every dimension we see considerable improvements. For instance, a considerable improvement in Pearson's $\rho$ is achieved for assessment of \textit{frame prediction quality} (`NSFrames' in Table \ref{tab:subtaskres}, +20.5\% $\rho$) and \textit{coreference quality} 
(`Reentrancies' in Table \ref{tab:subtaskres}, +18.5\%). 

A substantial error reduction is achieved in \textit{polarity} (`Negations', Table \ref{tab:subtaskres}), where we reduce the RMSE of the estimated F1 score by -8.6\%. When rating the SRL-quality of an AMR parse, our model reduces the RMSE by appr.\ 4\%. In general, improvements are obtained over almost all tested quality dimensions, both in RMSE reduction and increased correlation with the gold scores.


\subsection{Analysis}

\paragraph{Effect of data debiasing} We want to study the effect of the data set cleaning steps by analyzing the performance of our method and the baseline on three different versions of the data, with respect to estimated Smatch scores. The three versions are (i) \textbf{\textsc{$\frac{0}{2}$} = \textsc{AmrQuality}}, which is the original data; (ii) \textbf{\textsc{$\frac{1}{2}$}}, which is the data after the random re-split and score correction; (iii) \textbf{\textsc{$\frac{2}{2}$} = \textsc{AmrQualityClean}} which is our main data after the final debiasing step (shallow structure debiasing) has been applied.

\begin{table}
    \centering
    \scalebox{0.75}{
    \begin{tabular}{llrrr|r}
       & & \multicolumn{3}{c}{Pearson's $\rho$} & error\\
       
       data & method & P & R &  F1 & RMSE (F1) \\
       \midrule
       $\frac{0}{2}$ & LG-LSTM  & 0.72 & 0.78 & 0.77 & 0.138 \\
        & LG-LSTM$_{+aux}$  & 0.74 & 0.79& 0.78 & 0.137\\
        & ours  & 0.75 & 0.80 & 0.79 & 0.133\\
        & ours$_{+aux}$  & \textbf{0.76} & \textbf{0.81} & \textbf{0.80} & \textbf{0.132}\\
        \midrule
       $\frac{1}{2}$ & LG-LSTM & 0.67 & 0.73 & 0.72 & 0.120\\
         & ours  &  \textbf{0.68} & \textbf{0.75} & \textbf{0.74}& \textbf{0.117}\\
         \midrule
       $\frac{2}{2}$& LG-LSTM  & 0.60 & 0.68 & 0.66& 0.130\\
        & ours  &  \textbf{0.62} & \textbf{0.72} & \textbf{0.70} & \textbf{0.128}\\
    
       \midrule
    \end{tabular}}
    \caption{Performance-effects of data debiasing steps. $_{+aux}$ indicates a model variant that is trained using auxiliary losses that incorporate information about the other AMR aspects in the training process (see Fn.\footref{fn:aux}).}
    \label{tab:debias}
\end{table}

The results are shown in Table \ref{tab:debias}. We can make three main observations: (i) from the first to the second debiasing step, the baseline and our model have in common that Pearson's $\rho$ and the error decrease. While we cannot exactly explain why $\rho$ decreases, it is somewhat in line with recent research that observed performance drops when data was re-split \cite{gorman-bedrick-2019-need}. On the other hand, the error decrease can be explained by the random re-split that balances the target scores. (ii) The second debiasing step leads to a decrease in $\rho$ and an increase in error, for both models. This indicates that we have successfully removed shallow biases from the data that can give away the parse's source. (iii) On all considered versions of the data, the method performs better than the baseline.

\paragraph{AMRs: telling the good from the bad} In this experiment, we want to see how well the model can discriminate between good and bad graphs. To this aim, we create a five-way classification task: graphs are assigned the label `very bad' (Smatch F1 $<$ 0.25), `bad' (0.25 $\geq$ Smatch F1 $<$ 0.5), `good' (0.5 $\geq$ Smatch F1 $<$ 0.75), `very good' (0.75 $\geq$ Smatch F1 $<$ 0.95) and `excellent' (Smatch F1 $\geq$0.95). Here, we do not retrain the models with a classification objective but convert the estimated Smatch F1 to the corresponding label. Since the classes are situated on a nominal scale, and ordinary classification metrics would not fully reflect the performance, we also use quadratic weighted kappa \cite{cohen1968weighted} for evaluation.

\begin{table}
    \centering
    \scalebox{0.73}{
    \begin{tabular}{l|rrrr}
    \toprule
                & majority & random & LG-LSTM & ours\\
                \midrule
         avg.\ F1 & 0.13 & 0.20 & 0.40 & 0.44$^{\dagger\ddagger}$  \\
         quadr.\ kappa & 0.0 & 0.03 &  0.53 & 0.60$^{\dagger\ddagger}$ \\
         \bottomrule
        
    \end{tabular}}
    \caption{Graph quality classification task. $\dagger$ ($\ddagger$) significance with paired t-test at p$<$0.05 (p$<$0.005) over 10 random inititalizations.}
    \label{tab:classperf}
\end{table}

The results are shown in Table \ref{tab:classperf}. All baselines, including LG-LSTM, are significantly outperformed by our approach, both in terms of macro F1 (+4 points, 10\% improvement) and quadratic kappa (+7 points, 13\% improvement).

\paragraph{How important is the dependency information?} To investigate this question, instead of feeding the dependency tree of the sentence, we only feed the sentence itself. To achieve this, we simply insert the tokens in the first row of the former dependency input image, and pad all remaining empty `pixels'. In this mode, the sentence encoding is similar to standard convolutional sentence encoders as they are typically used in many tasks \citep{kim-2014-convolutional}.

\begin{table}
    \centering
   \scalebox{0.75}{\begin{tabular}{llrrr}
    \toprule
       &Quality Dim.\  & LG-LSTM & ours & ours (no dep.) \\
       \midrule
       \parbox[t]{2mm}{\multirow{3}{*}{\rotatebox[origin=c]{90}{P's $\rho$}}} &Smatch F1 & 0.662$^{\pm 0.00}$ & 0.696$^{\pm 0.00}$ &  0.682$^{\pm 0.01}$\\ 
       &Smatch precision & 0.600$^{\pm 0.00}$ & 0.623$^{\pm 0.01}$ & 0.614$^{\pm 0.01}$\\
       & Smatch recall & 0.676$^{\pm 0.00}$ & 0.719$^{\pm 0.00}$ & 0.702$^{\pm 0.01}$\\
       \midrule
       \midrule
    \parbox[t]{2mm}{\multirow{3}{*}{\rotatebox[origin=c]{90}{RMSE}}} &Smatch F1 & 0.130$^{\pm 0.00}$ & 0.128$^{\pm 0.00}$ & 0.128$^{\pm 0.00}$ \\
       &Smatch precision  &  0.126$^{\pm 0.00}$ & 0.126$^{\pm 0.00}$ & 0.129$^{\pm 0.00}$\\
       &Smatch recall&  0.142$^{\pm 0.00}$ & 0.136$^{\pm 0.00}$ & 0.139$^{\pm 0.00}$\\
       \bottomrule
    \end{tabular}}
    \caption{Right column: results of our system when we abstain from feeding the dependency tree, and only show the sentence together with the candidate AMR.}
    \label{tab:sentonly}
\end{table}

The results are shown in the right column of Table \ref{tab:sentonly}. The performance drops are small but consistent across all analyzed dimensions, both in terms of error (0 to 2.2\% increase) and Pearson's $\rho$ (1.4 to 2.4\% decrease). This indicates that the dependency trees contain information that can be exploited by our model to better judge the AMR quality. We hypothesize that this is due to similarities between relations such as \textit{subj}/\textit{obj} (syntactic) or arg0/arg1 (semantic), etc. Yet, we see that this simpler model, which does not see the dependency tree, still outperforms the baseline, except in estimated precision, where the error is increased by 2.4\%.

\paragraph{Efficiency analysis}\label{sec:eff} Recently, in many countries, there have been efforts to reduce energy consumption and carbon emission. Since deep learning typically requires intensive GPU computing, this aspect is of increasing importance to researchers and applicants \citep{DBLP:conf/acl/StrubellGM19,Tang:2019:IGD:3307772.3328315,8697354}. To investigate energy consumption of our method and previous work, we monitor their GPU usage during training, assessing the following quantities : (i) avg.\ time per epoch, (ii) avg. watts GPU usage, (iii) kilowatts per epoch (in kWh). 
\begin{table}
    \centering
    \scalebox{0.65}{
    \begin{tabular}{|l|rr|rr|}
        \toprule
       GPU type &  \multicolumn{2}{c|}{GTX Titan} &  \multicolumn{2}{c|}{GTX 1080}\\
       \midrule
        
         method &LG-LSTM & ours&LG-LSTM & ours  \\
         \midrule
         avg.\ ep.\ time & 722s & 59s & 1582s & 64s\\
         avg.\ W & 105 & 166 & 45 & 128 \\
         kWh per epoch & 0.021 & 0.003 & 0.020 &  0.002 \\
         \bottomrule

    \end{tabular}}
    \caption{Efficiency analysis of two approaches.}
    \label{tab:eff}
\end{table}

\begin{figure}
    \centering
    \includegraphics[width=0.5\linewidth]{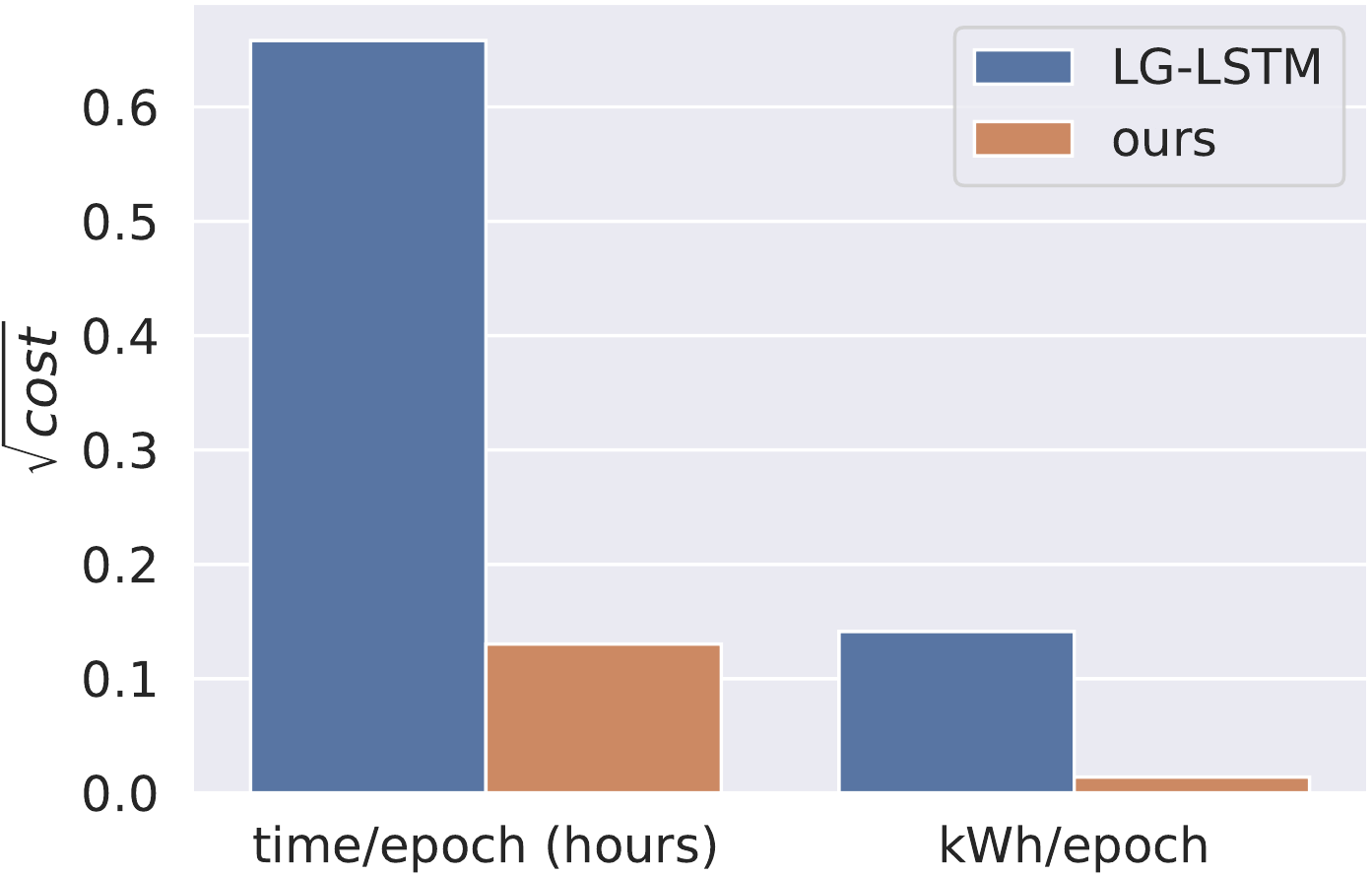}
    \caption{Training cost diagram of two approaches.}
    \label{fig:costred}
\end{figure}

The results of this analysis are displayed in Table \ref{tab:eff} and outlined in Figure \ref{fig:costred}. Our method consumes approximately 6.6 times less total kWh on a GTX Titan (10 times less on a GTX 1080). Directly related, it also reduces the training time: prior work requires appr. 1500s training time per epoch (GTX 1080), while our method requires appr.\ 60s per epoch (GTX 1080). The main reason for this is that our model does not depend on recurrent operations and profits more from parallelism. 

\section{Related work}\label{sec:rw}

\paragraph{Quality measurement of structured predictions} Since evaluating structured representations against human annotations is costly, systems have been developed that attempt an automatic quality assessment of these structures. Due to its popularity, much work has been conducted in \textit{machine translation} (MT) under the umbrella of \textit{quality estimation} (QE). QE can take place either on a word-level \citep{martins2017pushing}, sentence-level \citep{Specia09estimatingthe}, or document-level \citep{docque}. The conference on Machine Translation (WMT) has a long-standing workshop and shared task-series on MT quality assessment \citep{WMT5bojar-etal-2013-findings,WMTXbojar-etal-2014-findings,WMT6bojar-etal-2015-findings,WMT7bojar-etal-2016-findings,WMT4bojar-etal-2017-findings,WMT2specia-etal-2018-findings,WMT3fonseca-etal-2019-findings}. Quality estimation for neural language generation has been investigated, i.a., by \citet{scarton-etal-2016-quality}, and recently by  \citet{dusek-etal-2019-automatic}, who design a model that jointly learns to rate and rank generations, or by \citet{zopf-2018-estimating}, who predicts pair-wise preferences for generated summaries.

Furthermore, automatic techniques for the quality assessment of syntactic parses have been proposed. For instance, \citet{ravi2008automatic} formulate the task as a single-variable regression problem to assess the quality of constituency trees. A major difference to our work is that they try to assess the performance of a \textit{single parser}, while we aim at a parser-agnostic setting where candidate parses stem from \textit{different parsers}. Similarly, \citet{kawahara2008learning} predict a binary label that reflects whether the tree-quality lies above a certain threshold (or not). When multiple candidate parses are available, tree ranking methods \cite{zhu-etal-2015-ranking, zhou-etal-2016-search} may also be interpreted as some form of parse quality assessment (see \citet{do-rehbein-2020-neural} for a recent overview). Compared with assessing the quality of (abstract) meaning representations, judging about syntactic trees perhaps is a conceptually slightly simpler task, since the syntactic graphs are  more directly grounded in the sentence\footnote{In dependency trees, nodes are words; in constituency trees, nodes are (labeled) phrases; in meaning representations, words or phrases may be projected to abstract semantic nodes, or they may be omitted.}, and therefore it may be easier to judge whether graph components are correct, redundant, missing, or false.

In comparison to MT, automatic quality assessment of meaning representations is insufficiently researched. \citet{opitz-frank-2019-automatic} propose an LSTM based model that performs a multi-variate quality analysis of AMRs (constituting the baseline which we compared against). We believe that quality estimation approaches may also prove valuable for other meaning representation formalisms (MRs), such as, e.g., discourse representations \citep{KampReyle:93,Kamp81,abzianidze-etal-2019-first} or universal semantic dependencies \citep{reisinger-etal-2015-semantic,stengeleskin2019universal}. For example, since the manual creation of MRs is a notoriously laborious task, automatic quality assessment tools could assist humans in the annotation process (e.g., by serving as a cheap annotation quality check or by filtering automatic parses in active learning).

\paragraph{AMR metrics} When a gold graph is available, it can be used to compute the canonical AMR metric Smatch \citep{cai-knight-2013-smatch} that assesses matching triples. Furthermore, \citet{damonte-etal-2017-incremental} have extended Smatch to inspect various aspects of AMR. In this work, we have shown that our model can predict the expected outcomes of these metrics in the absence of the gold graph. Recently, more AMR metrics have been proposed, for example the Bleu-based \citep{papineni-etal-2002-bleu} SemBleu metric \citep{song-gildea-2019-sembleu}, Sema \cite{anchieta2019sema} or S$^2$match \citep{opitz2020amr}, a variant of Smatch. We plan to extend our model such that it also predicts these metrics.

\paragraph{AMR parsing} Recent advances in AMR parsing have been achieved by parsers that either predict latent alignments jointly with nodes \citep{lyu-titov-2018-amr}, or by transducing a graph from a sequence with a minimum spanning tree (MST) decoding algorithm \citep{zhang-etal-2019-amr}, or by focusing on core semantics in a top-down fashion \citep{cai-lam-2019-core}, or by performing auto-regressive decoding with a graph encoder \cite{cai2020amr}. Other approaches apply statistical machine translation \citep{pust-etal-2015-parsing} or sequence-to-sequence models, which tend to suffer from data scarcity issues and need considerable amounts of silver data to improve results \citep{NoordBos2017,konstas-etal-2017-neural}. Previously, alignment-based pipeline models have proved effective  \citep{flanigan-etal-2014-discriminative} or transition-based approaches that convert dependency trees step-by-step to AMR graphs \citep{wang-etal-2015-transition,wang2016camr,lindemann2020fast}.

\section{Conclusion}
 In this work, we have developed an approach to rate the quality of AMR graphs in the absence of costly gold data. Our model imitates a human judge that is confronted, `on paper', with the AMR in its native multi-line Penman format. We saw how this setup allowed efficient AMR processing with convolutions. Our experiments indicate that the method rates AMR quality more accurately and more efficiently than previous work.  
 
 \section*{Acknowledgments}

I am grateful to the anonymous reviewers for their valuable thoughts and comments. Moreover, I am grateful to Anette Frank for her thoughtful feedback on an earlier draft of this paper and her general guidance throughout my studies.

\bibliography{main}
\bibliographystyle{acl_natbib}
\typeout{get arXiv to do 4 passes: Label(s) may have changed. Rerun}
\end{document}